\documentclass[conference]{IEEEtran}
\IEEEoverridecommandlockouts
\usepackage{cite}
\usepackage{amsmath,amssymb,amsfonts}
\usepackage{algorithmic}
\usepackage{graphicx}
\usepackage{textcomp}
\usepackage{xcolor}

\usepackage[caption=false, font=footnotesize]{subfig}

\usepackage[capitalise]{cleveref} 
\usepackage{multirow} 
\usepackage{soul} 

\def\BibTeX{{\rm B\kern-.05em{\sc i\kern-.025em b}\kern-.08em
    T\kern-.1667em\lower.7ex\hbox{E}\kern-.125emX}}
\begin{document}

\title{BeSound: Bluetooth-Based Position Estimation Enhancing with Cross-Modality Distillation}

\IEEEaftertitletext{\vspace{-3\baselineskip}}
\author{\IEEEauthorblockN{Hymalai Bello}
\IEEEauthorblockA{\textit{DFKI and RPTU} \\
Kaiserslautern, Germany} \\
\and
\IEEEauthorblockN{Sungho Suh}
\IEEEauthorblockA{\textit{DFKI and RPTU} \\
Kaiserslautern, Germany }\\
\and
\IEEEauthorblockN{Bo Zhou}
\IEEEauthorblockA{\textit{DFKI and RPTU}  \\
Kaiserslautern, Germany }\\
\and
\IEEEauthorblockN{Paul Lukowicz}
\IEEEauthorblockA{\textit{DFKI and RPTU}  \\
Kaiserslautern, Germany} \\

}

\maketitle

\begin{abstract}
Smart factories leverage advanced technologies to optimize manufacturing processes and enhance efficiency. 
Implementing worker tracking systems, primarily through camera-based methods, ensures accurate monitoring. 
However, concerns about worker privacy and technology protection make it necessary to explore alternative approaches.  
We propose a non-visual, scalable solution using Bluetooth Low Energy (BLE) and ultrasound coordinates. 
BLE position estimation offers a very low-power and cost-effective solution, as the
technology is available on smartphones and is scalable due to the large number of smartphone users, facilitating worker localization and safety protocol transmission.
Ultrasound signals provide faster response times and higher accuracy but require custom hardware, increasing costs.
To combine the benefits of both modalities, we employ knowledge distillation (KD) from ultrasound signals to BLE RSSI data. 
Once the student model is trained, the model only takes as inputs the BLE-RSSI data for inference, retaining the advantages of ubiquity and low cost of BLE RSSI.
We tested our approach using data from an experiment with twelve participants in a smart factory test bed environment. 
We obtained an increase of 11.79\% in the F1-score compared to the baseline (target model without KD and trained with BLE-RSSI data only). 

\end{abstract}


\vspace{-10pt}
\section{Introduction}
\label{sec:Intro}
Nowadays, factories have been transformed into smart workplaces with automation at various levels. 
Using technologies to optimize manufacturing processes and improve overall efficiency. 
And focusing on human-centered artificial intelligence rather than technology-driven implementations to ensure a safe and inspiring environment for workers. 
Hence, human activity recognition and positioning systems are being developed as part of the framework to enable continuous monitoring of human behaviors. 
Tracking the workers' position is a relevant safety task. 
For example, on a petrochemical factory floor, chemical asphyxiation can provoke unconsciousness or death through suffocation. 
Some chemicals vaporize and mix with the air without showing any warning properties that raise the risk of oxygen deficiency \cite{islam2022smart}.
Thus, knowing the worker's position will increase the chances of the worker's survival by speeding up the arrival of the rescue team. 

For indoor personnel localization, vision-based solutions are the most accurate \cite{yang2020improved}.
For example, in \cite{maeda2023fusion}, a fusion between a vision-based, and radio-based real-time location systems (RTLS) for person identification is presented. 
The authors simulated the scenarios with images generated by a 3D game engine. 
And they used YOLOv3 for the object detection model combined with RSSI-based localization.  
They obtained a representative Rank-1 accuracy of 88.7\% with 4 people for 2 seconds window. 
A review of recent advances in vision-based indoor navigation is available in \cite{khan2022recent}. 
However, concerns about worker privacy and technology protection are relevant aspects to be considered. 
This led to alternative approaches. 
The state-of-the-art non-camera-based methods fall into four categories: dead reckoning technology, optical methods, radio-frequency-based (RF) methods, and sound-based localization systems. 

\textbf{Dead Reckoning} is based on tracking the user/asset by estimating the current position in a cumulative manner, starting from a known position. 
It is not dependent on external infrastructure, is low cost, has low latency, and can provide real-time tracking as it continuously updates the position based on data from previous movements. 
Due to cumulative prediction, its error also accumulates over time. 
This means that small errors in the measurement can lead to significant inaccuracies in the estimated future positions. 
It is also sensitive to sensor calibration and initialization error of the starting known position. 
Hence, this technology is fused with other modalities in the literature\cite{liu2017fusing,lee2012efficient,tsai1998localization}.

\textbf{Optical Methods}, usually emit ultra-short light pulses and measure the time of return to the source \cite{velten2012recovering,velten2013femto}. 
These methods use time-accurate equipment and high-power laser beams that could exceed the eye safety limit \cite{lindell2019wave}. 
The main drawback of these methods is their sensitivity to environmental distribution and lighting conditions.

\textbf{Radio Signals} and light waves are electromagnetic in origin and do not need a medium to propagate. 
They differ in their radiation frequency (energy) and wavelength. 
Radio-frequency signal has wavelengths at least 5000 times longer than visible light.  
This means that the RF signals experienced less diffusion when they encountered an obstacle/wall compared to light pulses\cite{yue2022cornerradar}. 
The authors in \cite{yue2022cornerradar}, used a radar-based solution to provide accurate around-corner indoor localization. 
The motivation is that it is pertinent to ensure that unmanned robots do not collide with factory personnel who suddenly appear around a corner.
Ultra-wideband (UWB) is an accurate RF-based tracking solution. 
Its limitation is mainly based on cost and the fact that it requires dedicated signal receivers.  
Moreover, RF technology such as WiFi and Bluetooth is highly scalable and low-cost. 
WiFi and BLE-based solutions are cost-effective, as the technology is available on smartphones and is scalable due to the large number of smartphone users.
The accuracy of such solutions is usually in meters. 
WiFi signals are more stable than BLE signals \cite{qureshi2019indoor}.
But, with proper signal processing, the BLE method can achieve results comparable WiFi method and require six times less power than WiFi signals \cite{qureshi2019indoor,ma2017ble,lindemann2016indoor}. 
On the other hand, BLE-tracking is sensitive to overcrowdedness\cite{BLEcrowds}.
The existence of human bodies is proven to influence Bluetooth signals. 
This can be used as an advantage if the purpose is to monitor crowds.  

\textbf{Ultrasonic Waves} are mechanical waves, requiring a medium to travel. 
The accuracy of ultrasonic sensors has proven to be in centimeters \cite{ijaz2013indoor}. 
In addition, they suffer less interference from obstacles and walls than WiFi/Bluetooth technology.  
They bounce off walls/objects unless the transducers are attached to the object \cite{roa2013ultrasonic}. 
This makes ultrasonic location systems reliable and less prone to interruptions.
However, it needs the installation of ultrasound transmitters and receivers in the environment (customized hardware), increasing the cost.

\begin{figure*}[ht]
    \centering
    \includegraphics[width=\textwidth]{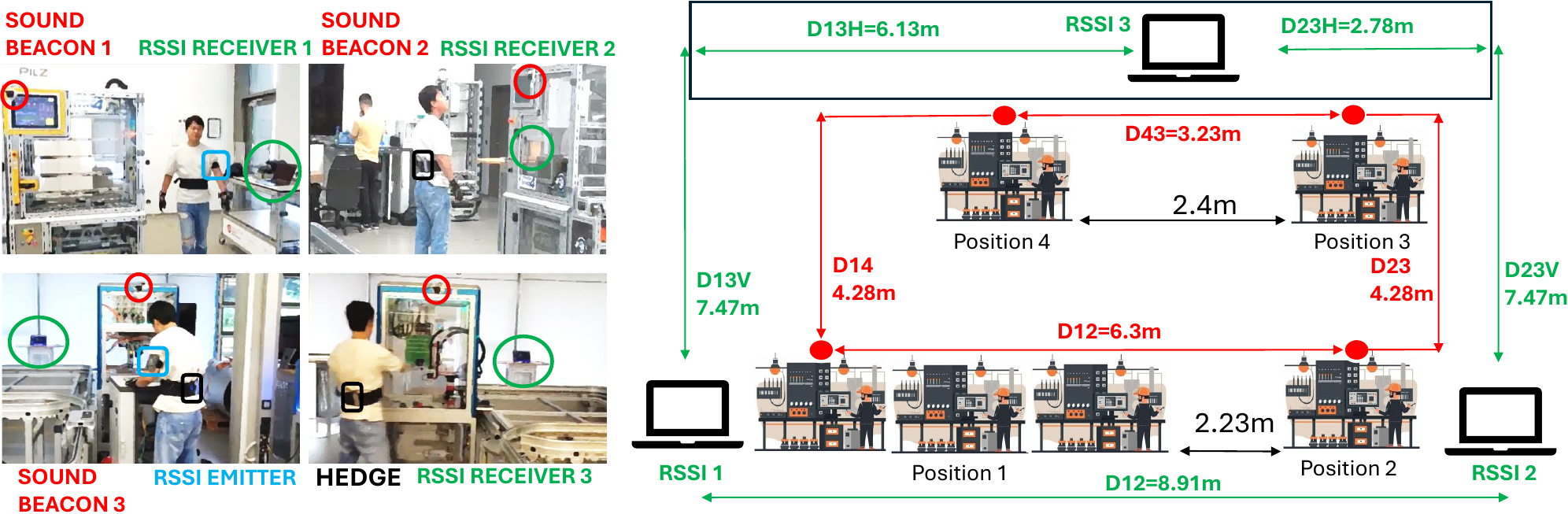}
    \caption{Experimental Setting Map. \textbf{Left} Volunteer Wearing the BLE-RSSI Emitter (Upper-Arm) and the Ultrasound Hedge (Lower back \textbf{BLACK BOX}) Together with the Spatial Position of The RSSI-Receivers in Green Color and the Position of the Ultrasound Beacons in Red Color. \textbf{Right} Top View Showing the Distances of the RSSI-Receivers and Ultrasound Beacon in the Smart Factory Test Bed.} 
    \label{fig:PositionalMap} 
    \vspace{-15pt}
\end{figure*}

In this work, we propose to enhance the low-cost and scalable option of BLE-RSSI-based localization with ultrasound coordinates.
We employ the teacher-to-student scheme. 
The teacher-to-student scheme is known as knowledge distillation\cite{hinton2015distilling}. 
The idea is to improve a smaller target model with a cumbersome teacher model's guidance. 
In our work, the teacher fused RSSI signals with ultrasound coordinates to leverage multimodal and multipositional information.
Then, the student (target model) is guided by the teacher. 
Once the student is trained, the position of the user is inferred by a model whose inputs are only RSSI channels.
The student is maintained only with RSSI signal dependency, which reduces cost, improves performance, and increases the ubiquity and scalability of the solution.
The \textbf{main contributions} of this work are described as follows: 
\begin{itemize}
    \item We present a solution based on knowledge distillation to enhance the low-cost and scalable localization method based on BLE-RSSI signals.
    The multimodal and multipositional teacher fuses BLE-RSSI signals with ultrasound coordinates and guides the student. 
    Once the student is trained, the user's position is inferred with only RSSI signals as inputs for full cross-modality transfer learning.
    \item We have conducted an experiment to track the position of twelve participants in a smart factory test bed. 
    In our approach, the BLE transmitter only needs to be enabled, and the receivers are distributed around the factory modules (see \cref{fig:PositionalMap}).
    Hence, no further application is deployed on the user's side. 
    The ultrasound data is collected at the same time as the BLE data for synchronized and fair data collection.
    \item We have evaluated the approach with three student models. 
    The students (target models) were designed with the same neural network structure but with a different number of parameters. 
    Experimental results show that with the smallest student with 754 parameters, the F1-score is 8.98\% higher than the target model baseline. 
    Moreover, for the biggest student with 4594 parameters, the F1-score is 70.96\%. 
    This represents a difference of 8.89\% in the F1-score compared to the teacher with 20416 parameters, and it is 11.79\% higher than the target model baseline.  
    These results were achieved by training the distilled students for 10 epochs. 
    In contrast to the target model's baselines trained for 100 epochs. 
\end{itemize}

The paper is organized as follows. 
\cref{sec:Related}, give an overview of the background and related work in BLE-based localization methods, ultrasound-based localization methods, and multimodal knowledge distillation. 
\cref{sec:Methods}, presents the hardware and experimental setting, and the BeSound knowledge distillation methodology. 
In \cref{sec:Results} are the results and discussions, including a list of limitations with an overview of future directions. 
Finally, \cref{sec:Conclusion} concludes the work.

\vspace{-10pt}
\section{Background and Related Work}
\label{sec:Related}
In this section, the motivation for the selected approach is expanded further and a background about the technology fused is given.
It is divided into three subsections; Bluetooth Low Energy-Based Indoor Localization Methods, Ultrasound-Based Indoor Localization Methods, and Multimodal Knowledge Distillation. 

\subsection{Bluetooth Low Energy-Based Indoor Localization Methods}
The BLE positioning system involves two elements; an anchor(s) and a tag(s).
Usually, BLE anchors also called beacons are positioned at fixed and known locations as emitters/receivers.  
The main idea is to estimate the tag(s) position, which is the emitter/receiver.
For an anchor-based localization system, several quantities can be measured to estimate the tag position.
The literature presents approaches based on the Angle of Arrival (AoA), Time of Flight (ToF), and Received Signal Strength Indicator (RSSI) \cite{milano2024ble,bahle2021using}. 

\textbf{The Angle of Arrival (AoA)} technology has been introduced to Bluetooth Low Energy (BLE) since Bluetooth 5.1 Core Specification to add directionality to BLE localization systems \cite{iakimenko2022bluetooth,mustafa2023high}. 
The AoA method intends to increase the location accuracy compared to the received signal strength indication (RSSI) approach. 
With the adoption of AoA, BLE localization systems can provide real-time indoor location, achieving an accuracy of 0.1 meters, which is 10 times higher than traditional indoor positioning technology (RSSI-based).
Commercially available options for this technology are BlueIOT (Beijing) Technology Co., Ltd \footnote{https://www.blueiot.com/}, and Dusun IoT \footnote{https://www.dusuniot.com/resources/technical-brief/the-benefits-of-ble-aoa-and-how-you-can-use-it-in-life/}. 
For an accuracy of less or equal to 0.1 meters, the technology requires customized hardware deployed in the environment as anchors and real-time location system (RTLS) tags \cite{cominelli2019dead}. 
Then the anchor with multiple antennas in the receiver calculates the AoA of the signal transmitted by the RTLS tag(s) using an RTLS location library. 
The RTLS library is in charge of processing the in-phase quadrature (IQ) samples received from the Bluetooth stack and implements multipath detection and azimuth and elevation calculation. 

\textbf{The ToF} method provides accuracy in the centimeter order \cite{giovanelli2018rssi, comuniello2021using}. 
The idea is to calculate the time taken by the electromagnetic wave to propagate from the transmitter to the receiver, which is proportional to the distance between them. 
The main conditions are; knowing the positions beforehand of at least three anchors for 2-dimensional tracking (4 anchors for 3D), a high-speed clock, and a common clock reference to be shared between the receiver and transmitter. 
For example, the propagation speed of an RF signal is around 300000 km/s. 
This means that to achieve a tracking accuracy of 30 cm the BLE radio and the timer employed for measuring the ToF should be clocked at 1 GHz. 

On the other hand, \textbf{the RSSI-based} localization technique is low-power, affordable, and simple to implement with standard hardware available in traditional versions of BLE \cite{milano2024ble}. 
Even though the accuracy is impacted by the line of sight, the system still captures an RSSI tag in a limited way. 
Moreover, it possesses an accuracy of more than one meter and is sensitive to crowded places\cite{BLEcrowds}. 
But still is considered a sufficient method for tracking personnel inside a building. 
Hence, different solutions have been proposed in the literature to improve the RSSI localization systems. 
The solutions include multichannel transmission using RSSI signal aggregation techniques, reducing the positioning error from 1.5 m to about 1 m.
The authors in \cite{giovanelli2018bluetooth} employ a Kalman filter to fuse RSSI and Time-of-Flight measurement data, and demonstrate the benefit of not relying only on RSSI, comparing ranging performed with and without the help of ToF. 
In \cite{zhu2022deep}, a fusion between AoA and RSSI with a two-step approach is introduced.
First, the AoA measurement is smoothed using a Kalman filter. 
Then a convolutional neural network (CNN) is used for feature extraction of RSSI and AoA, respectively. 
The fusion is performed at the feature level by a concatenating operation.  
Moreover, in \cite{koutris2022deep}, the authors present a deep neural network-based method that uses the RSSI and the in-phase quadrature (IQ) components of the BLE signals received at multiple anchors to estimate the AoA. 
The AoA is then used to achieve a localization accuracy of 70 cm. 

Overall, the BLE and RSSI-based localization method offers reduced complexity, low power, and scalability. 
It is a cost-effective solution, as the technology is available on smartphones and is scalable due to the large number of smartphone users.
Moreover, in our approach the movable tag (e.g. user's smartphone) needs only to be enabled, no further application is deployed on the user's side. 
The receivers, which are located on the factory modules capture the RSSI signal from the tag without even being paired to the BLE client. 

\subsection{Ultrasound-Based Indoor Localization Methods}
Ultrasound-based tracking has two elements; a hedge and the beacons. 
The hedge(s) is then tracked by the ultrasound beacons. 
Where the beacons' position is previously known. 
Ultrasound RTLS is often used as ground truth in the literature \cite{bian2021induced,amsters2019evaluation}. 
It provides high accuracy in location tracking, often achieving centimeter-level precision. 
The ultrasound wave propagates at 340 m/s in the air which is almost 1 million times slower than the BLE signal (RF wave). 
In theory, this indicates that it is 1,000,000 easier to achieve higher accuracy with ultrasound-based RTLS than with radio waves, such as ultra-wideband (UWB) based RTLS. 
Compared to WiFi/Bluetooth technologies, ultrasound technology typically experiences lower interference from obstacles, walls, or other electronic devices. 
This makes ultrasound RTLS reliable and less prone to disruptions. 
However, compared to WiFi/Bluetooth-based localization, the ultrasound method requires additional infrastructure costs. 
It needs the installation of ultrasound transmitters and receivers.
The need for a higher density of transmitters in certain environments will increase the complexity and cost of the entire system. 
All of the above indicates that ultrasonic system accuracy (centimeter-level) is desirable in RTLS. 
Still, the simplicity, cost-effectiveness, and scalability of a solution such as BLE-RSSI localization are also desirable in a positioning system. 

\subsection{Multimodal Knowledge Distillation}
The large-scale deep neural network models have been successful in many fields, achieving high performance at the cost of huge computational complexity and massive storage requirements. 
They possess the ability to learn and maneuver billions of model parameters to discover non-linear patterns in the data. 
Despite their success, it is cumbersome to train them and deploy them on devices with limited resources such as smartphones and microcontroller-based devices. 
With this in mind, many researchers have developed model compression and acceleration techniques. 
Knowledge distillation is one of those techniques. 
The idea is to learn a small model, called the student with the help/guidance of a larger model, called the teacher \cite{gou2021knowledge}. 
Knowledge distillation leverages the reduced size of the student and at some level the performance of the teacher. 
Lightweight models offer real-time deployment in embedded devices. 
This quality is particularly recommendable for RTLS. 

The theory behind the technique is established in \cite{buciluǎ2006model,hinton2015distilling}. 
The authors present a vanilla version of the technique. 
But before explaining it, a little background is needed. 
In a classification problem, a neural network produces class probabilities by using a ``softmax" output layer. 
This layer converts the logit computed for each class into a probability by comparing it with the other logit values. 
In this conversion exists a term called temperature ``T". 
``T" is normally set to 1, but using a larger number for ``T" will produce a softer probability distribution over classes. 
The idea is to transfer knowledge from a big model, the teacher, with a high temperature in its softmax. 
Then the performance of a small distilled model, the student, is improved by using a weighted average of two different objective functions. 
The first objective function is the cross-entropy with the soft targets. 
This cross-entropy is computed using the same high temperature in the softmax of the distilled model as was used for generating the soft targets from the teacher model. 
The second objective function is the cross-entropy with the correct labels (hard targets) and a temperature of 1.
For the weight of the objective functions, the authors found that the best results were generally obtained by using a considerably lower weight on the second objective function. 
A detail to consider is that the magnitudes of the gradients produced by the soft targets scale as $\frac{1}{T^2}$ it is important to multiply them by $T^2$ when using both hard and soft targets. 
In the ideal case, after applying knowledge distillation the output will be a smaller and capable model for the task at hand. 

The teacher-student approach can also be used to transfer learning from one sensing modality to another. 
In \cite{liang2022audioimu}, the authors augmented an IMU-based model with audio signals. 
Instead of training with only IMU data, an advanced audio-based teacher guided the training.  
The technique is capable of transferring rich context information from the audio-based model to the inertial-based model. 
There was an increment of 4.4\% compared to the baseline (inertial model alone) for recognizing 23 activities of daily living. 
Once the student is trained, the human activity recognition (HAR) model only takes as input the motion data for inference. 
In \cite{patidar2023vax}, the authors use the same methodology of cross-modality transfer learning, which they call VAX. 
VAX means video/audio to X, where the video/audio is the teacher modality and X is a privacy-sensitive sensor.
The motivation is that video/audio models are widely spread and trained with a large corpus of labeled training data. 
While privacy preserving modalities such as mmWave Doppler radar and inertial measurement units (IMUs) do not readily generalize across environments and require significant in-situ training data. 
The author's idea is to alleviate the lack of trained labeled data in privacy-aware sensor-based models with the guidance of the video/audio teacher. 
They achieved an improvement of 5\% average accuracy compared to the student baseline (from 79\% to 84\%). 

In this work, we propose to enhance the RSSI localization technique by distilling the knowledge of a multimodal teacher. 
The teacher fused RSSI signals with ultrasound coordinates to leverage multimodal information. 
Then, the student is guided by the multimodal teacher. 
Once the student is trained, the position of the user is inferred by a model whose inputs are only RSSI channels. 
The student is kept only with RSSI signal dependency, reducing cost, improving performance, and increasing the ubiquity and scalability of the solution.

\vspace{-10pt}
\section{Method}
\label{sec:Methods}
\subsection{Hardware and Experimental Setting}
The system consists of three parts and is distributed as shown in \cref{fig:PositionalMap}. 
The first part is a BLE tag attached to the volunteer's upper arm. 
The second part is the RSSI receivers deployed around the factory modules. 
The reason for the selection of dedicated boards for the BLE tag and BLE receivers is that we recorded additional sensing modalities not available in smartphones, in addition to those presented in this paper. 
But ultimately, the scalability of our method depends on the possibility of using smartphones for the final application. 
And the third part is the ultrasound tracking system. 
The map's setting aims to recognize in which module the worker is working (Positions 1-4).

\textit{The BLE tag} consists of the development board Adafruit Feather Sense. 
The Adafruit board was attached to the participant's upper arm with a smartphone Sport armband. 
Inside the feather sense, the BLE communication is handled by the Nordic nRF52832. 
It is a general-purpose multiprotocol SoC. 
The nRF52832 supports Bluetooth Low Energy, including the high-speed 2 Mbps feature. 
Moreover, Bluetooth mesh can be run concurrently with Bluetooth LE, enabling smartphones to provision, commission, configure and control mesh nodes. 
This means the device can be configured to interact with multiple nodes in a mesh network through a publish/subscribe paradigm. 
The RSSI resolution is 1 dB and the maximum transmission power is +4dBm (selected one). 

\textit{The RSSI receivers} were three Mackbook-Pro deployed around the smart factory modules, running a Python RSSI reception script. 
The specifications of the receivers are in \cref{tab:ReceiversInfo}. 
The position of the receivers is depicted in \cref{fig:PositionalMap} in green color. 
The sampling rate of the RSSI signal is around 50 Hz. 

\textit{The ultrasound localization system} is the off-the-shelf system from Marvelmind Robotics. 
The version of the Marvelmind navigation package was 1.0.8 (2018). 
The system provides indoor positioning and navigation data with a precision of 2 cm.
Ultrasound-based tracking has two elements; a hedge and the beacons. 
The hedge is then tracked by the ultrasound beacons. 
Where the beacons' position is previously known. 
In our setting, the hedge is attached to the volunteer's back using a lifting belt. 
And, the ultrasound beacons were on the smart factory modules as shown in \cref{fig:PositionalMap} in red color. 
The Marvelmind system used the information from the beacons and tracked the hedge in a three-coordinate space (X, Y, and Z). 
The sampling rate for the position is around 10 Hz. 
 
\begin{table}[!t]
\footnotesize
    \centering
    \caption{BLE-RSSI Receivers Specifications. All the Receivers were MacBook Pros from Different Models. }
    \begin{tabular}{c c c c c}
        \hline
         Model& OS & Position & Bluetooth Version\\
        \hline
         A2141-2019& Sonoma 14.3 & 1& 5.0\\
         A1502-2014 & Big Sur 11.7.7 & 2 &4.0 \\
         A1502-2015& Monterrey 12.7.2 & 3 &4.0\\
         \hline
    \end{tabular}
    \label{tab:ReceiversInfo}
    \vspace{-15pt}
\end{table}

\subsection{Experimental Procedure}
Twelve volunteers were recruited.
They identify themselves as ten male and two female. 
Their age ranges from 23 to 59 years old (mean of 30.75). 
The height ranges from 160-184 centimeters (mean 178 cm). 
Only one of the participants was left-handed. 
The participants wore the RSSI-emitter (tag) on the left upper arm. 
The ultrasound hedge is placed on the volunteer's back using a lifting belt.
At the beginning of the experiment, the sensors' data is synchronized in front of a video camera. 
The camera time is then used as a global clock to synchronize the data from the RSSI receivers and ultrasound beacons. 
The volunteers were asked to walk around the modules and simulate working activities on a factory floor. 
The activities performed by each volunteer were categorized as walking, touching the screen/buttons, opening the door, working inside the module, and closing the door. 
Each of these activities was performed on each module in each session.
A total of five sessions per participant were recorded. 
In between every session, the hardware was removed from the wearer and a period of ten to twenty minutes rest was enforced.
This makes the results accountable for the re-wearing of the system, which is typically expected in wearable devices. 
Each session lasted around 20 minutes on average.
One participant performed two sessions one day and three sessions another. 
One volunteer only performs three sessions in total. 
And one volunteer has only four valid sessions. 
The participants with less than five sessions were only included in the training set (2 participants).
For ten participants, the data is split into 4 sessions for training and 1 session for testing. 
A 5-fold cross-validation with a leave-one-session-out evaluation scheme is performed. \footnote{All participants signed an agreement following the policies of the university's committee for protecting human subjects and following the Declaration of Helsinki \cite{Helsinki1975}. }

\subsection{BeSound Knowledge Distillation}
\begin{figure*}[ht]
    \centering
    \includegraphics[width=0.9\textwidth]{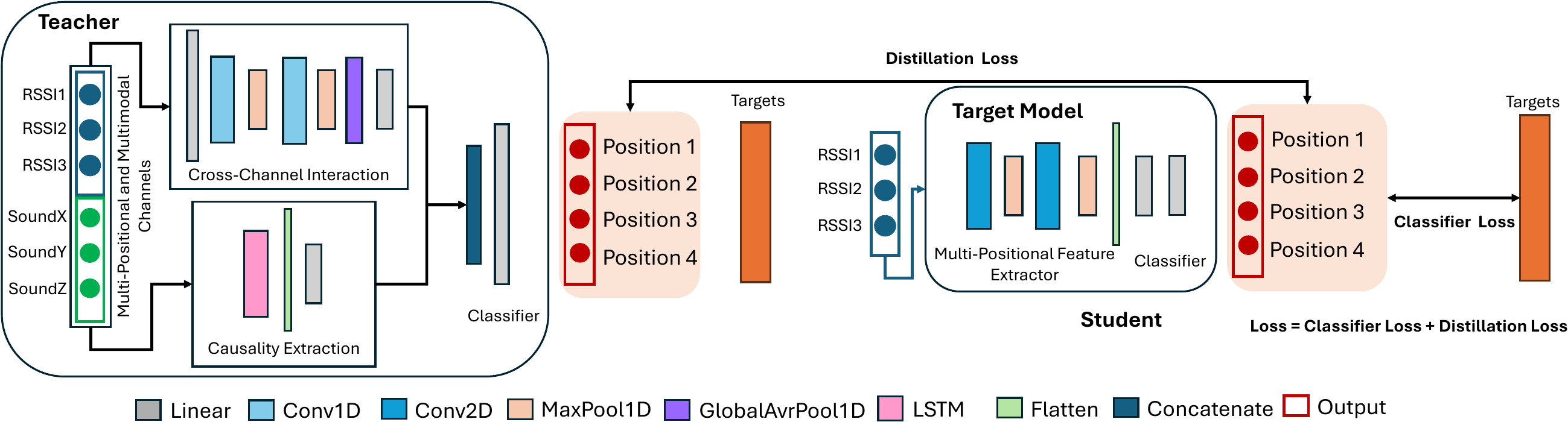}
    \caption{BeSound Knowledge Distillation Approach. \textbf{Left} The Multimodal and Multipositional Teacher Consists of Two Concatenate Networks; One For Cross-Channel Interaction Feature Extraction and One Network for Causality Extraction (LSTM-Based). \textbf{Right} The Student Consists of a Multipositional Feature Extractor and a Classifier. The Distillation is Applied at the Logit Level to Improve the Position Estimation of the Student.}
    \label{fig:KDArchitecture}
    \vspace{-15pt}
\end{figure*}

The first step of the algorithm is the signal pre-processing. 
We obtained three BLE-RSSI signals from the BLE receivers and coordinates X, Y, and Z from the Marvelmind Ultrasound system. 
For the RSSI values a DC offset removal is performed by session (mean subtraction by channel). 
The subtraction of the mean is intended to highlight signal variations rather than absolute values. 
The absolute values of the RSSI BLE signal vary even under static conditions, but the variations are comparatively stable.
Therefore, a small variation indicates a stable position, and a high variation indicates relevant changes in the user's position.
This helps the model to train faster, obtaining compatible results between sessions.
This is followed by a 2-second resample window (100 samples at 50Hz). 
Then, a Butterworth low pass filter of 3 Hz was used to remove the jitter on the RSSI signal. 
For the case of ultrasound-based coordinates a resample to 2 seconds window is performed. 
This was followed by a Butterworth low pass filter of 10 Hz to remove the ringing from the resampling procedure and keep the pattern of the original sampling rate of 10 Hz. 
The resampling to 50Hz is for synchronization purposes with the video-ground truth with 50 frames per second. 
The data is manually annotated according to the position of the participants in the recorded video.
A sliding window of 2 seconds with a step size of 0.5 seconds is employed.
After pre-processing, the data set structure is six channels with a window size of 2 seconds and 25\% overlapping. 
The ground truth of the worker's position is extracted from the recorded videos manually. 
The categories were four positions related to the work modules depicted in \cref{fig:PositionalMap}. 
The next step in our approach is to train the teacher for the knowledge distillation to the student. 
The TensorFlow 2.15.0 version is used to train the networks. 
The evaluation scheme is defined as a 5-fold cross-validation with leave-one-session out.  
The training of teacher and student baselines ran for 100 epochs with early stopping (patience 20) and restoring the best weight option enabled to avoid overfitting. 
The training for the distilled model ran for 10 epochs. 
And, a 64-batch size was selected. 

\textbf{The Teacher:} In this work, we have trained a multimodal and multipositional teacher model. 
RSSI-BLE signals from the three receivers and the X, Y, and Z coordinates of the ultrasound localization system were fused.
This fusion is considered multimodal and multipositional due to the sensor's distribution on the experimental map in \cref{fig:PositionalMap}.
The structure of the teacher model is depicted in \cref{fig:KDArchitecture} \textbf{Left}. 
The input layer consists of six channels, three BLE-RSSI, and three ultrasound position coordinates. 
These six channels are fed to two separate neural networks (NNs). 
One of the neural networks is focused on extracting features of the cross-channel interaction between the modalities and positions.
The details of the NN for cross-channel interaction extraction are in \cref{table:CrossChannel}.
And the second neural network aims to extract the causality of the multimodal time series input.
Both networks are then concatenated and fed into a classifier layer (Linear layer), followed by an output layer with the softmax activation function.  
The causality extractor network is based on Long-Short-Term Memory (LSTM)\cite{hochreiter1997long}, and the details are in \cref{table:Causality}. 
The combination of the two concatenate NNs can capture spatial and temporal information simultaneously, thus 
effectively solving complex time series problems. 
With this structure, the teacher has 4594 parameters (79.75 KB). 
The optimizer is Adam with a learning rate of 0.001. 
The loss function is categorical cross-entropy and the metric to monitor is accuracy. 

\begin{table}[!t]
\caption{Implementation Details of the Neural Network for Cross-Channel Interaction of the Teacher Model} 
\centering
\resizebox{0.6\columnwidth}{!}{
\begin{tabular}{l l}
\hline
Layers & Configuration\\
\hline 
\multirow{2}{*}{Dense} & Size 10    \\
                       & Activation Function ReLu   \\
\hline
\multirow{6}{*}{Conv1D-1} & Filters 40 \\
                                & Kernel 10\\
                                & Padding "Same"\\
                                & Activation Function ReLu\\
                                & MaxPooling Size 10\\
                                & Dropout 0.5\\
\hline
\multirow{6}{*}{Conv1D-2}    & Filters 40 \\
                                & Kernel 10\\
                                & Padding "Same"\\
                                & Activation Function ReLu\\
                                & MaxPooling Size 10\\
                                & GlobalAvgPooling \\
\hline
\multirow{2}{*}{Dense}     & Size 10  \\
                            & Activation Function ReLu\\

\hline
\end{tabular}}
\label{table:CrossChannel}
\vspace{-15pt}
\end{table}

\begin{table}[!t]
\centering
\caption{Implementation Details of the Neural Network for Causality Extraction of the Teacher Model} 
\resizebox{0.5\columnwidth}{!}{
\begin{tabular}{l l}
\hline
Layers & Configuration\\
\hline 
\multirow{3}{*}{LSTM} &Units 16    \\
                       &Return Sequences True \\
                       &Flatten\\
\hline
\multirow{2}{*}{Dense} &Size 10 \\
                       &Activation Function ReLu\\

\hline
\end{tabular}}
\label{table:Causality}
\vspace{-15pt}
\end{table}

\textbf{The Target Model:}
The idea is to explore the performance enhancement of small and simple networks that can be deployed in wearable and embedded devices with a reduced impact on power consumption and memory. 
The structure of the student/target NN is depicted in \cref{fig:KDArchitecture} \textbf{Right}.
The input layer consists of three BLE-RSSI channels. 
The network combines a multipositional feature extraction and a classifier with two linear layers. 
The NN is a CNN-based model. 
Three sizes of student models were evaluated. 
The sizes are defined by the number of filters and kernels of the CNN layers. 
The number of parameters for each student is 754, 1824, and 4594, respectively. 
Which is around 27.07, 11.19, and 4.44 times smaller than the teacher with 20416 parameters, respectively.
The details of the networks are in \cref{table:StudentDetails}
The optimizer is Adadelta with a learning rate of 0.9. 
The loss function is categorical cross-entropy and the metric to monitor is accuracy.

\begin{table}[!t]
\caption{Implementation Details of the Neural Network of the Student/Target Model} 
\centering
\resizebox{\columnwidth}{!}{
\begin{tabular}{c c c c}
\hline
Layers & Student/Target 1 & Student/Target 2 & Student/Target 3 \\
\hline 
\multirow{6}{*}{Conv2D-1} &Filters 5 & Filters 10 & Filters 20 \\
                       &Kernel 3 & Kernel 5 & Kernel 5\\
                       &Padding "Same" & Padding "Same" & Padding "Same"\\
                       &ReLu &ReLu &ReLu\\
                       &MaxPool 5 &MaxPool 5 &MaxPool 5\\
                       &Dropout 0.2&Dropout 0.2 &Dropout 0.2\\
\hline
\multirow{6}{*}{Conv2D-2} &Filters 5 & Filters 10 & Filters 20 \\
                       &Kernel 3 & Kernel 5 & Kernel 5\\
                       &Padding "Same" & Padding "Same" & Padding "Same"\\
                       &ReLu &ReLu &ReLu\\
                       &MaxPool 5 &MaxPool 5 &MaxPool 5\\
                       &Flatten&Flatten &Flatten\\
\hline
\multirow{3}{*}{Dense} &Size 10 & Size 10 & Size 10 \\
                       &ReLu &ReLu &ReLu\\
                       &Dropout 0.1&Dropout 0.1 &Dropout 0.1\\

\hline
\end{tabular}}
\label{table:StudentDetails}
\vspace{-15pt}
\end{table}

\textbf{Distilled Student:}
In this work, the vanilla version of knowledge distillation is applied \cite{hinton2015distilling}. 
The optimizer is Adam \cite{kingma2014adam} with a learning rate of 0.001. 
The distilled students were trained with the goal of minimizing the student loss and the distillation loss. 
The equation for the total loss is in \cref{eq:loss}. 
Where $\alpha$ (set to 0.1) is the weight for the objective function, $sl$ is the student loss and $dl$ is the distillation loss.
\begin{equation}
    loss = \alpha \times sl + (1 - \alpha) \times dl
    \label{eq:loss}
\end{equation}
The student loss is defined as sparse categorical cross-entropy loss with logit enabled. 
The distillation loss is defined in \cref{eq:Dloss}. 
Where $T$ is the distillation temperature, set to 10. 
$KLD$ is the Kullback-Leibler divergence loss (KLD). 
KLD is a non-symmetric metric that measures the relative entropy or the difference in information represented by two distributions.
It measures the distance between two data distributions to show how far they are from each other.
This metric is typically used in knowledge distillation tasks \cite{kim2021comparing}. 
\begin{equation}
    dl = KLD(V_t, V_p)* T^2
    \label{eq:Dloss}
\end{equation}
The true values of KLD ($V_t$) are defined as the softmax function in \cref{eq:Softmax}.  
Where $\Vec{z}$ is the input vector defined as $\frac{T_p}{T}$, where $T_p$ is the teacher prediction and $T$ is the distillation temperature. 
$e^{z_{i}}$ is the exponential of input vector, $e^{z_{j}}$ is the exponential of the output vector and $K$ is the number of categories. 
The predicted values of KLD ($V_p$) are defined as the softmax function in \cref{eq:Softmax}. 
Where $\Vec{z}$ is defined as $\frac{S_p}{T}$, where $S_p$ is the student prediction and $T$ is the distillation temperature. 
\begin{equation}
    \sigma(\Vec{z}) = \frac{e^{z_{i}}}{\sum_{j=1}^K e^{z_{j}}} \ \ \ for\ i=1,2,3,\dots K
    \label{eq:Softmax}
\end{equation}

\section{Result and Discussion}
\label{sec:Results}
\begin{figure}[!t]
    \centering
    \includegraphics[width=\columnwidth]{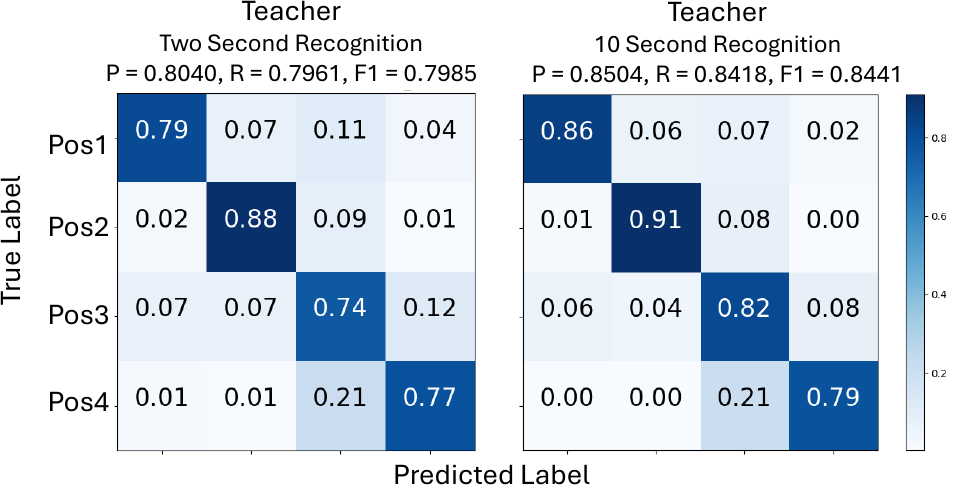}
    \caption{Multimodal and Multipositional Teacher Average Results with 5Fold-Cross Validation with Leave-Session-Out Evaluation Scheme. \textbf{Left} Confusion Matrix Result for a Window Size of Two Seconds with F1-Score of 79.85\%. \textbf{Right} Confusion Matrix Result for a Window Size of Ten Seconds with F1-Score of 84.41\%.}
    \label{fig:TeacherResults}
    \vspace{-15pt}
\end{figure}

\cref{fig:TeacherResults} presents the results of the trained teacher. 
An F1-score of 79.85\% and 84.41\% is obtained for the two-second and ten-second window recognition, respectively.
The highest confusion pair is Position 3 and Position 4. 
This might be because of our experimental setting map (see \cref{fig:PositionalMap}). 
For the case of BLE-RSSI signals, Position 3 and Position 4 are covered by only one RSSI receiver. 
Unlike Position 1 and Position 3, both have their own RSSI receiver.
In the case of the ultrasound beacons, we observed that volunteers to moved between Position 3 and Position 4 tended to select the path behind the factory modules, where the ultrasound antennas were not pointing.
When a 10-second recognition window is selected, the results improve by 4.56\%, indicating that the detection modalities need time to stabilize after the participant moves between modules. 

\begin{figure*}[ht]
    \centering
    \subfloat[Unimodal and Multipositional Target Model Baselines Average Results with 5Fold-Cross Validation with Leave-One-Session Out Evaluation Scheme. \textbf{Target Model 1 (754 Parameters)} Results for a Window Size of Two and Ten Seconds with F1-Score of 58.10\% and 60.19\%, respectively. \textbf{Target Model 2 (1824 Parameters)} Results for a Window Size of Two and Ten Seconds with F1-Score of 59.48\% and 61.91\%, respectively. \textbf{Target Model 3 (4594 Parameters)} Results for a Window Size of Two and Ten Seconds with F1-Score of 59.10\% and 62.31\%, respectively.]{\includegraphics[width=\textwidth]{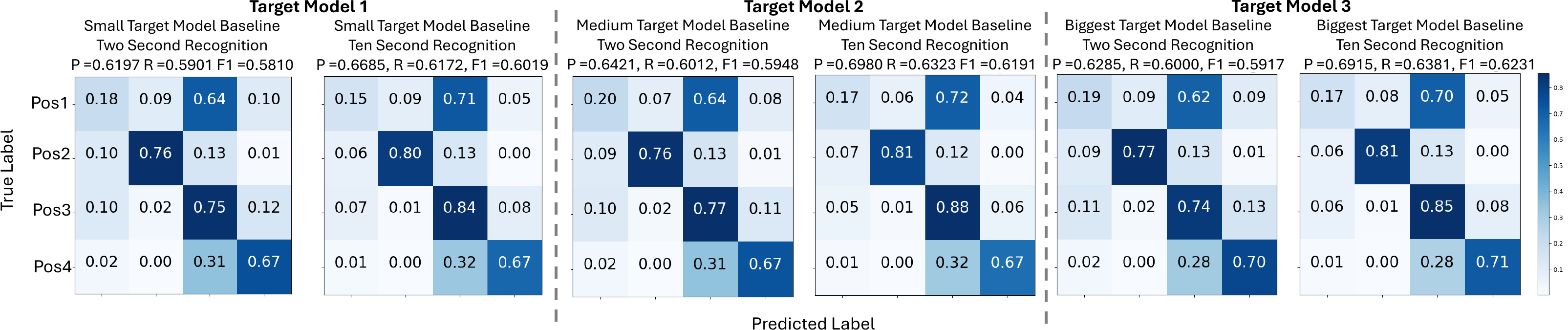}}
    \hfill    
    \subfloat[Distilled Student Models Average Results with 5Fold-Cross Validation with Leave-One-Session Out Evaluation Scheme. \textbf{Distilled Student 1} Results for a Window Size of Two and Ten Seconds with F1-Score of 67.08\% and 72.84\%, respectively. \textbf{Distilled Student 2} Results for a Window Size of Two and Ten Seconds with F1-Score of 67.79\% and 73.17\%, respectively. \textbf{Distilled Student 3} Results for a Window Size of Two and Ten Seconds with F1-Score of 70.96\% and 76.04\%, respectively.]{\includegraphics[width=\textwidth]{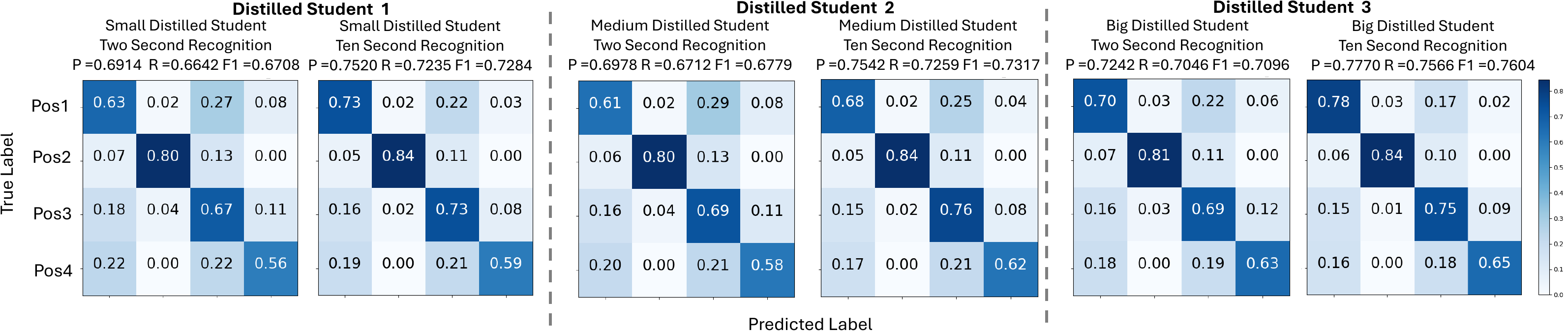}}
    \caption{Target Models and Distilled Students Models Average Results with 5Fold-Cross Validation with Leave-One-Session Out Evaluation Scheme.}
    \label{fig:Results}
    \vspace{-15pt}
\end{figure*}

\cref{fig:Results} (a) depicts the confusion matrices for the target models' baselines to be later compared with the distilled student models.
The differences between the students/targets are the number of filters and kernels in the CNN layers. 
The first student has only 754 parameters, 27 times smaller than the teacher (20416 parameters).
And, the biggest student has 4594 parameters, which is 4 times smaller than the teacher model. 
All the target models' baselines present the highest misclassification when estimating Position 1. 
The Position 1 as shown in \cref{fig:PositionalMap} is 3 times bigger than the rest of the factory modules. 
BLE-RSSI localization is not suitable for long-range positioning. 
BLE technology has been developed for short-range communications up to 10 meters. 
The 10-meter range is without considering interference from surrounding obstacles/people. 
Thus, determining Position 1 can be challenging with only BLE-RSSI signals as input. 

\cref{fig:Results} (b) presents the confusion matrices of the distilled students. 
For the case of Distilled Student 1 an F1-score of 67.08 \% is achieved. 
This result indicates an 8.98\% performance increase compared to the target model 1 baseline with 58.10\%. 
The student is 27 times smaller than the teacher and it can achieve an F1-score difference of 12.77\% compared to the teacher (79.85\% F1-score).
The medium-size model, Distilled Student 2, has a performance of 12.06\% F1-score lower than the teacher and 8.31\% higher than the target model 2 baseline. 
The Distilled Student 3 outperformed the target model baseline by 11.79\% F1-score, and the difference between the student and the teacher was reduced to 8.89\%. 
This model is 4 times smaller than the teacher.
The recall of Position 1 is highly increased for all the distilled models compared to the target model baselines (see \cref{fig:Results}). 
Increasing the number of parameters of the student from 754 to 4594 (16\% increment) has a 3.88\% increase in the F1-score.  

Overall, our results show the potential of cross-modal knowledge distillation for localization systems. 
All distilled student models outperformed the target model baselines. 
This increase in performance indicates that the multimodal and multipositional teacher effectively guides the student. 
With this method, the model is compressed and improved. 
Moreover, in the inference, only the BLE-RSSI signals are the inputs. 
This means that our solution retains the ubiquity and low-cost advantages of BLE-RSSI and, at the same time, improves performance.

On the other hand, our solution has several \textbf{limitations} and possibilities for improvement.
This is not a complete list but they are relevant limitations of our work. 
\begin{itemize}
    \item \textbf{Detection Radio:} 
    The detection radio is dependent on the number of receivers for the wearable BLE RSSI. 
    In this case, three RSSI receiver devices were deployed (one per module) in the environment.
    In the future, it will be interesting to remove the dependency on sensors on the environment. 
    This could be achieved by distilling cross-modal knowledge to a learner whose inference is only based on IMU data.
    \item  \textbf{Experimental Setting:}
    The experimental setting map in \cref{fig:PositionalMap} shows a minimum distance between the factory modules of 2 meters. 
    The method will increase in relevance if it is possible to capture distance with an accuracy of centimeters. 
    The experiment is performed by one volunteer at a time with a maximum of 8 people around and a minimum of 2 people around. 
    Hence, additional challenges might arise if the positions are estimated in crowded spaces. 
    \item \textbf{Temperature Value and Weight for Objective Function:} 
    We have selected a temperature value of 10 and a weight of 0.1 for the objective function of the knowledge distillation. 
    This selection follows typical values of those parameters for the vanilla version of distillation. 
    But, it might be valuable to test different values and analyze the performance results. 
    \item  \textbf{Kullback-Leibler Divergence Loss (KLD):}
    We employ the KLD loss because this loss has achieved considerable success by controlling the ``soft" targets via the temperature scaling parameter. 
    In \cite{kim2021comparing}, the authors proved that the MSE loss outperforms the KLD loss, explained by the difference in the penultimate layer representations between the two losses.
    \item \textbf{Knowledge Distillation at Logit Level:}
    The method we have used is based on the distillation loss feedback in the output layer. 
    This can be extended to the feature layer in the student network. 
    Moreover, there exist other methods such as self-distillation \cite{zhang2019your}, learning from multiple teachers \cite{you2017learning}, and combining knowledge distillation with data-free compression \cite{lopes2017data}, among others. 
    \item \textbf{Smartphone Antennas Variability:} In our experiment design we purposely employed Apple products for the BLE receivers due to their stability in receiving the BLE RSSI. 
    However, we have used different versions of MacBooks to add variability in the receiving antennas, which is to be expected in the real-world scenario where there is a huge variety of smartphone vendors. 
    The variability included in the experiment may not be sufficient to represent the variety of smartphone antenna designs, so in the future, it is pertinent to quantify the impact of the hardware on the signal reception. 
\end{itemize}

\vspace{-5pt}
\section{Conclusion}
\label{sec:Conclusion}
\vspace{-5pt}
In this paper, we present a framework for improving a localization system based on BLE-RSSI signals. 
We have introduced multimodal and multipositional BLE-RSSI and ultrasonic coordinates information using a teacher-student framework. 
The BLE-RSSI student model benefits from teacher guidance during training, while relying solely on BLE-RSSI data for inference. 
This approach retains the advantages of ubiquity, low cost, and scalability of the BLE-RSSI positioning method. 
Based on a study with 12 participants in which they simulated working and moving between modules of the smart factory test bed, we show the extent to which our framework improved performance. 
We have evaluated the approach with three student/target models. 
The students were designed with the same neural network structure but with a different number of parameters.
Experimental results show that with the smallest student with 754 parameters (27 times smaller than the teacher), the F1-score is 8.98\% higher than the target model baseline. 
Moreover, for the biggest student with 4594 parameters (4 times smaller than the teacher), the F1 score is 11.79\% higher than the target model baseline.
These results were achieved by training the distilled students for 10 epochs. 
In contrast to the target model's baselines trained for 100 epochs.  
This makes our approach efficient and sustainable with fewer epochs per training and higher performance. 
We also discussed a list of relevant limitations of our approach with future solutions. 

\vspace{-10pt}
\section*{Acknowledgment}
The research reported in this paper was partially supported by the German Federal Ministry of Education and Research (BMBF) in the projects SocialWear (01IW20002) and VidGenSense (01IW21003).
We would also like to thank the volunteers and the student help of Luis Sanchez.  

\bibliographystyle{IEEEtran}

\bibliography{References}

\end{document}